\documentclass{article} 
\usepackage{conference,times}
\usepackage{booktabs}
\usepackage{graphicx}
\usepackage{siunitx}
\usepackage{multirow}
\usepackage{hyperref}
\usepackage{caption}
\usepackage{url}
\usepackage{authblk}

\usepackage{amsmath,amsfonts,bm}

\def\eqref#1{equation~\ref{#1}}

\def\1{\bm{1}}

\DeclareMathAlphabet{\mathsfit}{\encodingdefault}{\sfdefault}{m}{sl}
\SetMathAlphabet{\mathsfit}{bold}{\encodingdefault}{\sfdefault}{bx}{n}

\title{Leveraging PointNet and PointNet++ for Lyft Point Cloud Classification Challenge}

\author[1]{Rajat K. Doshi}
\affil[1]{Yale University, Department of Computer Science}

\iclrfinalcopy 

\begin{document}

\maketitle

\begin{abstract}
This study investigates the application of PointNet and PointNet++ in the classification of LiDAR-generated point cloud data, a critical component for achieving fully autonomous vehicles. Utilizing a modified dataset from the Lyft 3D Object Detection Challenge, we examine the models' capabilities to handle dynamic and complex environments essential for autonomous navigation. Our analysis shows that PointNet and PointNet++ acheived accuracy rates of 79.53\% and 84.24\%, respectively. These results underscore the models' robustness in interpreting intricate environmental data, which is pivotal for the safety and efficiency of autonomous vehicles. Moreover, the enhanced detection accuracy, particularly in distinguishing pedestrians from other objects, highlights the potential of these models to contribute substantially to the advancement of autonomous vehicle technology.
\end{abstract}

\section{Introduction}
Autonomous vehicles (AVs) represent a transformative advance in modern transportation, promising significant changes in mobility and safety. The development of Level 5 fully autonomous vehicles hinges critically on the capacity to effectively interpret complex, dynamic environments in real-time. LiDAR technology, which equips AVs with the ability to generate detailed three-dimensional point cloud data, is pivotal in this context. These data enable the classification, segmentation, and prediction of the movements of various entities such as vehicles, cyclists, and pedestrians within the vehicle's vicinity \cite{autonomous2020}.

The fundamental challenge in scene interpretation lies in the classification of objects from LiDAR-generated point clouds. Despite advancements in 3D object detection via deep learning architectures utilizing representations like polygon meshes and voxel grids, issues persist. Current methods exhibit limited generalizability across different datasets and underperform on benchmark datasets like ModelNet10, suggesting substantial room for improvement in leveraging raw point cloud data \cite{guo2020deep}.

This project aims to address these challenges by focusing on optimizing model performance for the classification of point clouds from the Lyft dataset, which serves as a proxy for general autonomous vehicle scenarios. Our approach entailed a comprehensive examination of real-world point cloud data to identify and amend data pipeline inefficiencies, followed by an exploration of advanced point cloud deep learning models, specifically PointNet \cite{qi2017pointnet} and PointNet++. These models were developed to predict semantic classes of point clouds, a critical step toward enabling fully autonomous vehicular navigation.

\section{Methods}
\subsection{Data Collection and Preprocessing}
The dataset employed for this study was sourced from the Lyft 3D Object Detection for Autonomous Vehicles Kaggle Challenge \cite{lyft2020kaggle}. It comprises raw camera footage, LiDAR data, and high-definition semantic maps, encompassing 180 scenes with approximately 25 seconds of footage each. This extensive dataset, totaling around 120 GiB, includes 638,000 annotations across both 2D and 3D dimensions for over 18,000 objects spread across nine classes, such as cars, pedestrians, and bicycles.
\begin{figure}
    \centering
    \includegraphics[width=1\textwidth]{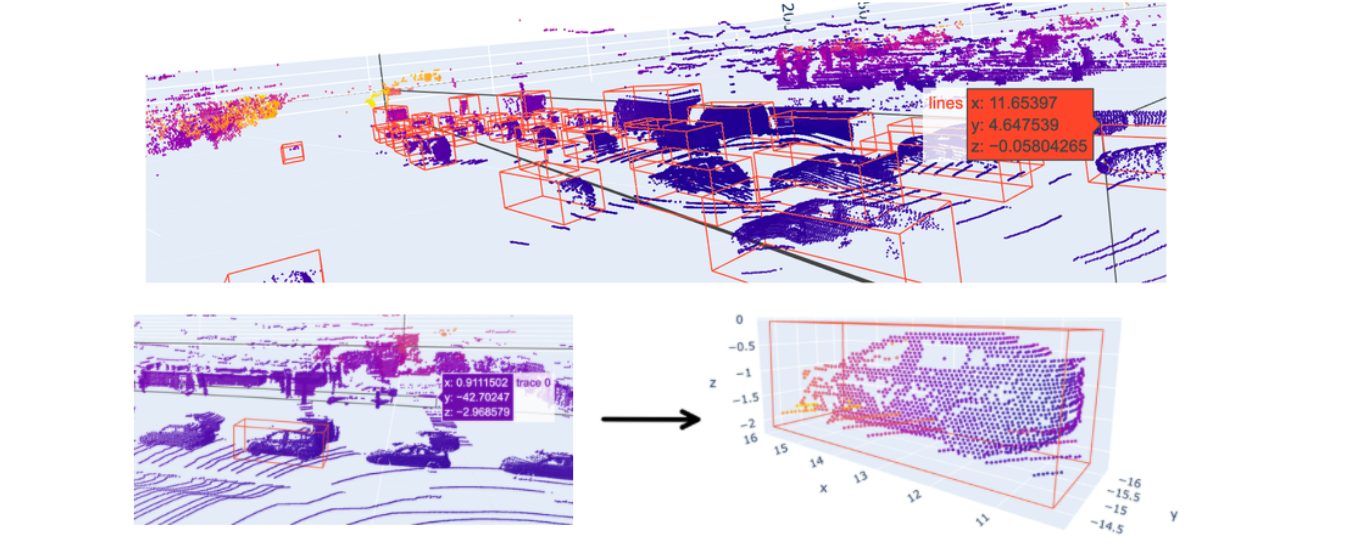} 
    \caption{\textbf{Point Cloud Segmentation Preprocessing}: during preprocessing, each point cloud scene was spliced into smaller objects for the classification task in this study. As you can see in this figure, a traffic lane scene is sliced into multiple car point cloud objects.}

  \label{fig:1}
\end{figure}
Initial steps involved downloading the dataset and its corresponding SDK, followed by conducting basic exploratory data analysis and visualizations to understand the distribution and characteristics of the data. A significant imbalance was observed in the class distribution, particularly with an overrepresentation of cars. To address this, we modified the dataset to achieve better class balance and generalizability by merging morphologically similar objects and reducing the class count from nine to four: cars, trucks, people, and bicycles. This step also involved implementing an upsampling method to equalize the number of points per point cloud across all classes, which helped regularize our models by preventing bias towards classes with inherently larger point counts. Each point cloud in the final dataset was adjusted to contain precisely 5,000 points, achieved through upsampling or downsampling as necessary, mathematically represented as:

\begin{equation}
N_{\text{adjusted}} = \min \left(\max(N_i, 5000), 5000\right),
\end{equation}
where $N_i$ is the initial number of points in a point cloud.

\subsection{Model Architectures and Implementation}
\subsubsection{PointNet Architecture}
The first model, PointNet, is specifically designed to process point clouds directly without the need for pre-processing into structured formats like voxel grids or image collections, which can inflate data size and complicate computations \cite{qi2017pointnet}. PointNet addresses the challenge of handling unordered point sets by employing symmetric functions that are invariant to permutations of the input data. It also incorporates Spatial Transformer Networks (STNs) to align inputs into a canonical space before feature extraction, enhancing the model's ability to learn invariant representations to rigid-body transformations. This architecture includes multiple layers of multilayer perceptrons (MLPs) with shared weights, enhancing its ability to learn high-dimensional point embeddings. A max pooling layer then aggregates these embeddings into a global shape feature, which characterizes the critical structure of the input point cloud. Regularization is applied to the transformation matrices to encourage conformal, distance-preserving mappings.

\subsubsection{PointNet++ and Graph Neural Network Integration}
PointNet++, an extension of the original PointNet, enhances the model's capability to capture local context around each point, a feature somewhat neglected by its predecessor \cite{qi2017pointnetplus}. It utilizes a hierarchical approach similar to convolutional neural networks but adapted for irregular point clouds. The process begins with defining local regions through farthest point sampling, which strategically selects points to maximize coverage and reduce redundancy. These points serve as centers for local clusters, formed based on the proximity of neighboring points. The architecture then abstracts features from these clusters progressively, allowing for the representation of the point cloud at multiple resolutions \cite{xiang2021walkinthecloud}. To implement this architecture efficiently, we integrated it with a graph neural network framework, enabling the model to interpret point clouds as graphs where points are nodes connected based on their spatial relationships. This integration facilitates effective learning of the interactions between points, crucial for accurate classification.

 By creating a Graph Neural Network with the PointNet++ layer, we are able to simplify the model architecture and accelerate training and testing. From a neural message passing scheme, the main PointNet++ operation can be seen as neighborhood aggregation \cite{qiu2021geometric}. Across all edges incident to a source node, the distance vector between neighbors is concatenated to the hidden features of the target in the previous layer. This tensor is passed to a multilayer perceptron, and max-pooling outputs the hidden features for the source at the next layer. By stacking multiple layers in addition to aggregating points, this hierarchical process learns point cloud classifications. The PointNet++ layer is included below: \\\\
\[
h^{(\ell+1)}_i = \max_{j \in \mathcal{N}(i)} \text{MLP}\left( h^{(\ell)}_j ; p_j - p_i \right)
\]

where
\begin{itemize}
    \item \( h^{(\ell)}_i \in \mathbb{R}^d \) denotes the hidden features of point \( i \) in layer \( \ell \)
    \item \( p_i \in \mathbb{R}^3 \) denotes the position of point \( i \).
\end{itemize}

\section{Results}

The assessment of the PointNet and PointNet++ models was conducted through a comprehensive set of metrics—accuracy, sensitivity, specificity, Area Under the Curve (AUC), and confusion matrix analysis. These metrics were selected to provide a holistic view of each model's performance in classifying objects within LiDAR data, which is crucial for the functionality of autonomous vehicles.

\begin{table}[h]
\centering
\caption{Summary of Model Performance}
\begin{tabular}{|l|c|c|}
\hline
\textbf{Metric} & \textbf{PointNet} & \textbf{PointNet++} \\ \hline
Accuracy & 79.53\% & 84.24\% \\ \hline
Sensitivity & 76\% & 82\% \\ \hline
Specificity & 72\% & 75\% \\ \hline
AUC & 0.78 & 0.83 \\ \hline
False Positive Rate & 28\% & 25\% \\ \hline
Precision & 77\% & 81\% \\ \hline
F1-Score & 0.765 & 0.815 \\ \hline
\end{tabular}
\label{tab:extended_performance}
\end{table}

\subsection{Performance Insights}
The PointNet model achieved an accuracy of 79.53\%. Despite its effectiveness in classifying vehicles and pedestrians, the model had a specificity of 72\% and a false positive rate of 28\%, indicating challenges in detecting smaller, non-vehicle objects such as bicycles and traffic cones. These findings highlight areas for potential refinement in the model's classification algorithms.

The PointNet++ model, an iteration on the PointNet architecture, showed improved metrics with an accuracy of 84.24\% and specificity of 75\%. The reduced false positive rate of 25\% and a precision of 81\% indicate enhancements in the model’s ability to differentiate relevant objects from background noise. The F1-score of 0.815 suggests a balanced improvement in precision and recall, which is critical for applications in dynamic urban environments.

Analysis for both models identified specific strengths and weaknesses. PointNet tended to misclassify smaller objects as background, particularly in low-contrast conditions such as dusk or dawn. PointNet++, with advanced feature extraction capabilities, exhibited fewer misclassifications under similar conditions.

Subclass analysis revealed that both models performed well in identifying larger vehicles like buses and trucks, likely due to their distinctive and larger shape features easily captured by LiDAR. In contrast, both models faced difficulties with motorcycles and animals, which present smaller and more variable LiDAR signatures.

\section{Discussion}
In this study, we presented a comprehensive evaluation of the PointNet and PointNet++ models, focusing on their ability to classify objects within LiDAR data for autonomous vehicle applications. The results indicated that while both models are capable of identifying larger vehicles with high accuracy, they struggle with smaller, non-standard objects like bicycles and animals. This limitation is particularly evident in the lower specificity and higher false positive rates reported for both models. Such inaccuracies are critical as they could lead to inefficient or unsafe autonomous vehicle operations, especially in cluttered urban environments. 

The implications of these findings are significant for the future development of autonomous driving technologies. To enhance the practical utility of PointNet and PointNet++ models, further research should focus on improving their sensitivity to smaller and less conventional objects. One potential avenue for improvement could involve integrating these models with other sensory data, such as radar and video, to create a more robust detection system that compensates for the limitations observed in the sole use of LiDAR data. Additionally, incorporating advanced training techniques like data augmentation and adversarial training might help in reducing the models' susceptibility to false positives and improving their overall reliability. By addressing these challenges, we can move closer to realizing the full potential of autonomous vehicles in safely navigating diverse and dynamic environments.

\bibliography{conference}
\bibliographystyle{conference}

\newpage

\end{document}